\pdfoutput=1

\documentclass[11pt]{article}

\usepackage{ACL2023}
\usepackage{adjustbox}

\usepackage{times}
\usepackage{latexsym}
\usepackage{multirow}
\usepackage{longtable}
\usepackage{lscape}     
\usepackage{graphicx}   

\usepackage[T1]{fontenc}

\usepackage[utf8]{inputenc}
\usepackage[table]{xcolor}

\usepackage{microtype}

\usepackage{inconsolata}

\usepackage{xcolor}
\usepackage{amsmath}

\title{RoBiologyDataChoiceQA: A Romanian Dataset for improving Biology understanding of Large Language Models}

\author{Dragoș-Dumitru Ghinea \and Adela-Nicoleta Corbeanu \and Marius-Adrian Dumitran \\
        University of Bucharest}

\begin{document}
\maketitle
\begin{abstract}

In recent years, large language models (LLMs) have demonstrated significant potential across various natural language processing (NLP) tasks. However, their performance in domain-specific applications and non-English languages remains less explored. This study introduces a novel Romanian-language dataset for multiple-choice biology questions, carefully curated to assess LLM comprehension and reasoning capabilities in scientific contexts. Containing approximately 14,000 questions, the dataset provides a comprehensive resource for evaluating and improving LLM performance in biology.

We benchmark several popular LLMs, analyzing their accuracy, reasoning patterns, and ability to understand domain-specific terminology and linguistic nuances. Additionally, we perform comprehensive experiments to evaluate the impact of prompt engineering, fine-tuning, and other optimization techniques on model performance. Our findings highlight both the strengths and limitations of current LLMs in handling specialized knowledge tasks in low-resource languages, offering valuable insights for future research and development.

\end{abstract}

\section{Introduction}

Large language models (LLMs) have achieved impressive results across a wide range of natural language processing (NLP) tasks. However, their performance often degrades in specialized domains and non-English languages, making Romania’s rich tradition in biology an ideal context for evaluating LLMs' scientific reasoning in a relatively low-resource setting.

To rigorously examine and ultimately improve LLM competence on such domain-specific tasks, we created a Romanian-language dataset of multiple-choice biology questions. The dataset was developed to assess and enhance LLM performance on authentic Romanian biology tests. It enables the evaluation of model accuracy in a realistic multiple-choice setting and can also be used to fine-tune LLMs on domain-specific Romanian biology terminology.

Our dataset comprises questions from two prestigious national sources: the Romanian Biology Olympiad and medical school admission examinations. The Olympiad is the country’s largest biology competition, targeting middle- and high-school students, while medical entrance exams rigorously assess pre-university candidates on foundational biological knowledge. Together, these sources provide a comprehensive and challenging collection of questions, covering a broad range of biological topics, difficulty levels, and linguistic complexity.

This study goes beyond simple benchmarking. We conduct extensive experiments to explore how various factors such as prompt engineering strategies, model origin, and domain-specific fine-tuning influence model performance. Our statistical analyses provide insights into how well LLMs grasp Romanian biological concepts, reveal common failure patterns, and highlight differences across model types.

Our contributions are threefold: (1) we introduce a carefully curated Romanian-language biology dataset suitable for benchmarking and domain adaptation; (2) we assess the capabilities of leading LLMs in scientific reasoning within a low-resource language setting, building on previous work that shows persistent challenges in this area \citep{huang-chang-2023-towards}; and (3) we present an in-depth analysis of performance variation across experimental conditions, offering insights that can inform future model development and deployment in specialized domains.

By presenting these findings, we aim to foster further research into LLM applications in non-English languages and domain-specific tasks, and to promote NLP advancements tailored to educational and scientific contexts. Our dataset and methodology support continued progress in this important area.

\section{Related work}

Biomedical question-answering (QA) datasets have played a crucial role in advancing domain-specific language models. PubMedQA \cite{jin2019pubmedqa} introduced a large-scale English-language biomedical QA dataset with 1,000 expert-annotated, 61,200 unlabeled, and 211,300 artificially generated \textit{yes/no/maybe} questions. While valuable for scientific text comprehension, it does not include multiple-choice questions, which require more complex reasoning over structured information.

A more relevant effort is MedQA \cite{jin2021medqa}, an open-domain multiple-choice QA dataset collected from professional medical board exams. MedQA covers three languages — English (12,723 questions), simplified Chinese (34,251 questions), and traditional Chinese (14,123 questions) — and requires models to select the correct answer from multiple options rather than extracting answers directly from text. Similarly, MedMCQA \cite{ankit2022medmcqa} is an English-language multiple-choice QA dataset designed for medical entrance exams, containing over 194,000 questions. Unlike MedQA, which focuses on board exam questions, MedMCQA emphasizes a wide range of medical knowledge, testing over ten different reasoning abilities.

Efforts to develop language models specialized for Romanian biology are quite limited. One notable contribution is RoQLlama, a lightweight Romanian-adapted language model designed to enhance NLP performance in Romanian-language applications \cite{dima-etal-2024-roqllama}. RoQLlama was evaluated using the RoMedQA dataset \cite{romedqa}, a specialized collection of Romanian medical school examination questions.

Our work surpasses this effort by introducing a carefully curated and extended Romanian-language biology dataset extracted from multiple sources, going beyond single-choice questions. We also fine-tune promising models and perform multiple benchmarks. Fine-tuning on our dataset significantly improves LLM performance, making it a valuable resource for enhancing language models in biology. By focusing on this domain, our dataset diversifies the range of available domain-specific resources for Romanian, complementing previous contributions in the medical field and aiming for deeper reasoning.

Guidance on creating and documenting high-quality NLP datasets is essential for ensuring the utility of research outcomes. The dataset documentation framework proposed by \citeauthor{Gebru2018DatasheetsFD}, \citeyear{Gebru2018DatasheetsFD} provided foundational insights for structuring the description and documentation of our dataset.

The use of LLMs in biology has shown significant potential for transforming research in the life sciences. \citeauthor{bhattacharya2023}, \citeyear{bhattacharya2023} explored the evolution of LLMs from textual comprehension tools to multimodal systems capable of analyzing complex biological data and contributing to advances in molecular biology and medicine. Their findings highlight the importance of LLMs in handling scientific reasoning and specialized terminology, which is central to our work.
\section{Dataset Composition}

\begin{figure}[h!]
  \centering
  \includegraphics[width=\columnwidth]{images/data-graphic.pdf}
  \caption{The data distribution based on question type and collection sources details.}
  \label{fig:data-graphic}
\end{figure}

\begin{figure}[h!]
  \centering
  \includegraphics[width=\columnwidth]{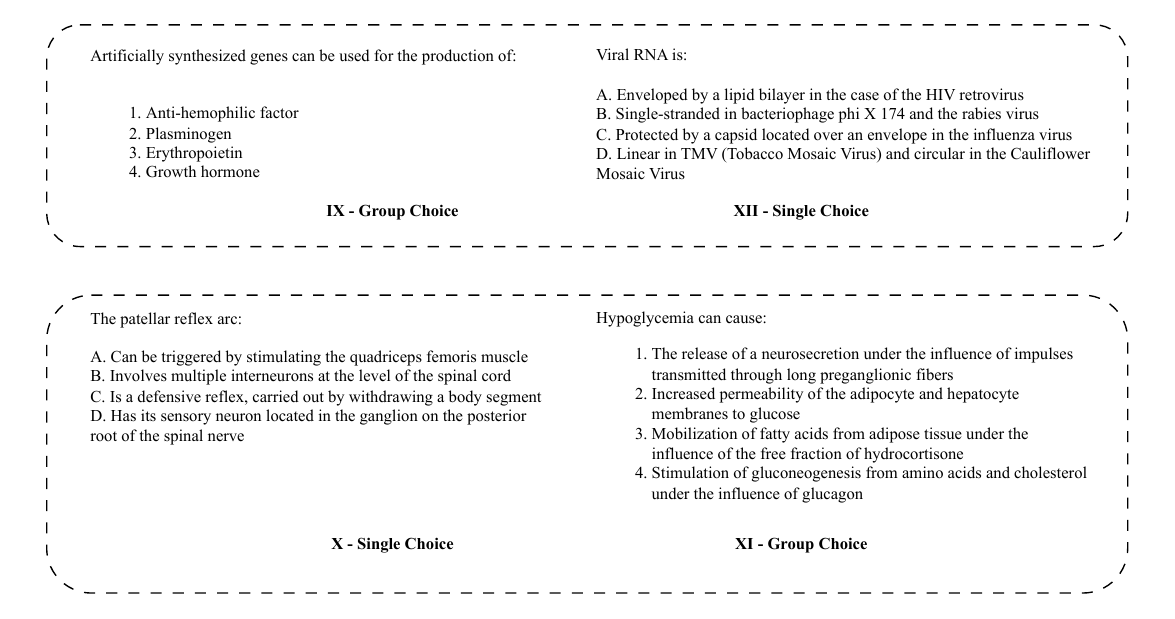}
  \caption{Examples of questions extracted and translated from the dataset}
  \label{fig:examples-by-grades}
\end{figure}

\subsection{Olympiads}

The \textit{Romanian National Biology Olympiad} is a multiple-choice-based competition structured in multiple stages, covering all high school grades and occasionally including middle school. A typical Olympiad exam consists of three primary question categories:

\begin{itemize}
    \item \textbf{Single-choice questions} – Typically, 30 questions with a single correct answer.
    \item \textbf{Group-choice questions} – Another 30 questions, where each answer can be one of five predefined lettered combinations (further details in \ref{sec:appendix_datasheet}).
    \item \textbf{Complex single-choice questions} – A set of 10 advanced problems requiring analytical problem-solving to determine the correct answer.
\end{itemize}

There are exceptions to this standard format, particularly in older exams or localized stages, where the structure may differ, featuring only single-choice questions or a varying number of items.

Olympiad data is collected exclusively from \textbf{PDF documents} available online, typically hosted on news websites, archived school portals, or dedicated Olympiad platforms such as \href{https://www.olimpiade.ro/}{olimpiade.ro}.

\begin{figure}[h!]
  \centering
  \includegraphics[width=\columnwidth]{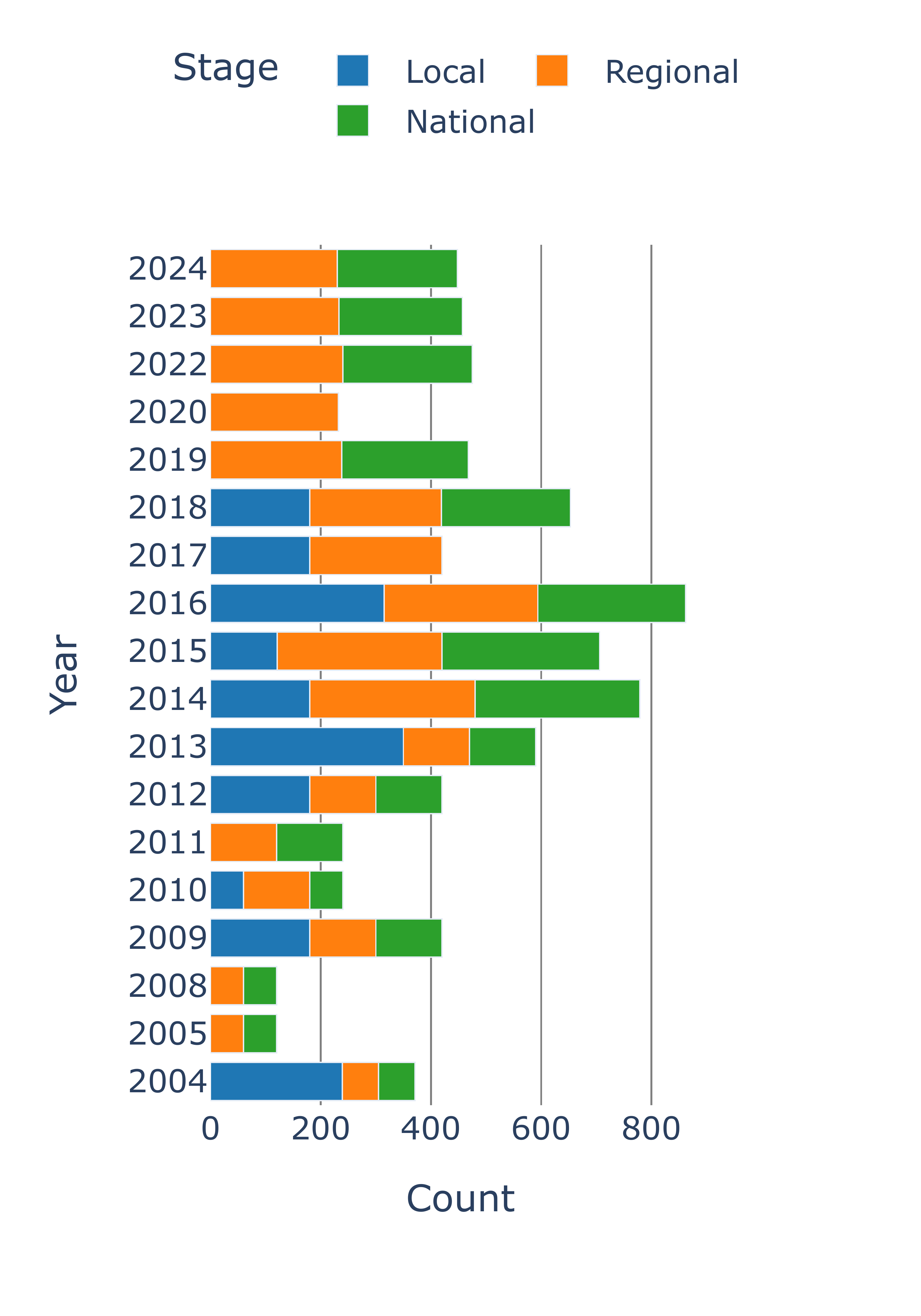}
  \caption{How many questions were collected from each year and of which type.}
  \label{fig:data-yearly-olympiads}
\end{figure}

As shown in Figure~\ref{fig:data-graphic}, we extract only \textbf{single-choice} and \textbf{group-choice} questions from multiple grades, covering various competition stages and years (Figure~\ref{fig:data-yearly-olympiads}). Given that the source documents are predominantly text-based PDFs (with occasional Word files, which we manually convert into PDFs), \textbf{PyMuPDF4LLM} \cite{pymupdf4llm} is used to extract content in Markdown format. The extracted text is subsequently parsed into question instances using \textbf{regular expressions}.

A major challenge in this process is \textbf{word fragmentation} due to inconsistencies in document formatting. To address this, we employ \textbf{Gemini 1.5 Flash} and \textbf{Gemma2 9B Instruct} for grammar correction, followed by manual validation. Despite instructions to preserve original meaning, models often altered the semantics, particularly by correcting intentionally wrong answer options. This suggests that LLMs exhibit a tendency to favor logically correct statements, indicating that they have either encountered similar data during training or have developed an implicit understanding of correctness through their learned representations.

\subsection{College Admission}

Several Romanian universities use \textbf{multiple-choice-based admission exams}, with each university providing a dedicated question book (\citeauthor{matusz2020biologie}, \citeyear{matusz2020biologie}; \citeauthor{pascu2020biologie}, \citeyear{pascu2020biologie}; \citeauthor{opincariu2018biologie}, \citeyear{opincariu2018biologie}). These books, authored by university professors, serve as the \textbf{primary study resource} for candidates, as the actual exam questions are guaranteed to be similar to them. Our dataset includes approximately \textbf{6,000 questions} collected from the admission preparation books of three universities (Figure~\ref{fig:data-graphic}).

Unlike the Olympiad materials, these documents are \textbf{scanned books in image-based PDFs}, necessitating Optical Character Recognition (OCR). The lack of Romanian-specialized OCR tools presents a challenge. While \textbf{docTR} \cite{liao2023doctr}
, a library known for strong English OCR performance, was tested, it proved inadequate for Romanian text. The most viable alternative was \textbf{Tesseract OCR}, optimized with \textbf{OpenCV-based noise removal preprocessing} \cite{kotwal2021optical}. However, this approach introduced challenges:
\\
\begin{itemize}
    \item \textbf{Inconsistent noise removal} – Some techniques improved OCR accuracy for one page while degrading performance on others.
    \item \textbf{Language constraints} – The texts, although in Romanian, contain \textbf{Greek letters} used for specialized terminology (e.g., $\alpha$, $\beta$, $\gamma$). While Tesseract supports multiple languages, enabling both Romanian and Greek led to \textbf{higher misinterpretation rates} rather than improved detection of Greek symbols.
\end{itemize}

To mitigate these issues, we explored \textbf{AI-based OCR solutions}, relying on context-aware processing for improved accuracy. The \textbf{Gemini Flash 1.5} model provided better results in recognizing text within scanned images. However, occasional hallucinations—such as \textbf{unintended duplication of questions}—necessitated \textbf{manual verification} to ensure proper extraction.

\subsection{Deduplication}

When identical questions with the same answer options appear across different tests or problem sets, we assign them a shared \texttt{dupe\_id}, a unique UUID identifying a group of duplicates. Each group contains at least two instances. A question is considered a duplicate if both its text and answer options match, regardless of option order, which, as a matter of fact, could impact performance \cite{pezeshkpour-hruschka-2024-large}. To detect slight rephrasings, we compare text embeddings generated with \textbf{jina-embeddings-v3} \cite{sturua2024jinaembeddingsv3multilingualembeddingstask}.

Rather than removing duplicates, we mark them, as it is unclear which instance should be deleted. Duplication data may also reveal relationships between different subjects. While duplicates remain in the dataset, users can filter them using the \texttt{dupe\_id} if needed. We ensure that no duplicates exist between the training, validation, and test splits to maintain dataset integrity.

\begin{figure}[h!]
  \centering
  \includegraphics[width=\columnwidth]{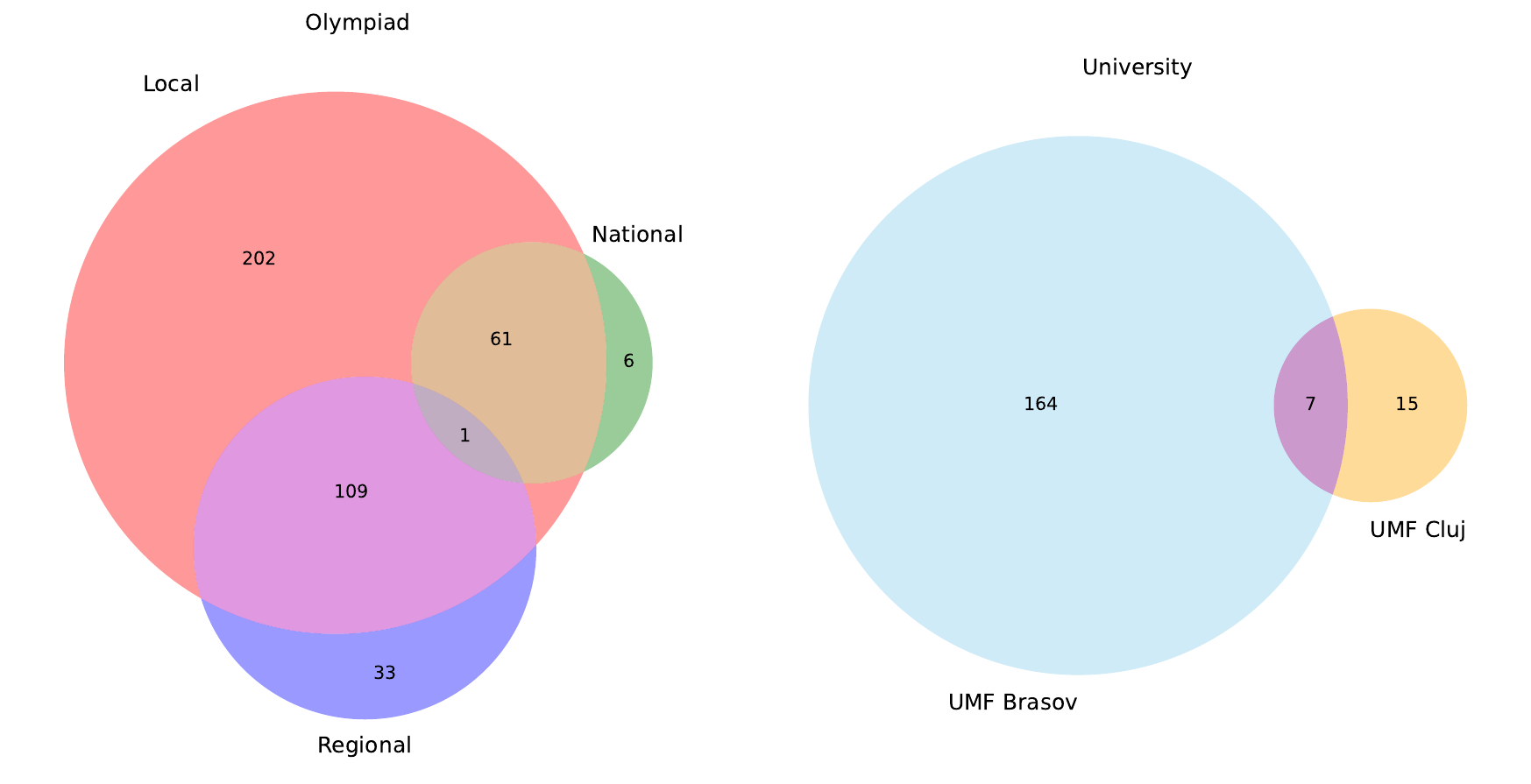}
  \caption{Duplication groups by stage. Overlaps indicate that the same question appears across all the participating stages. There is no duplicate question to be present in both olympiad and university subjects at the same time.}
  \label{fig:data-graphic}
\end{figure}

\subsection{Data Splits}

The dataset is split into 11,347 training, 1,374 validation, and 1,388 test questions. Stratified sampling was applied across grades, difficulty tiers (national, regional, local), and institutional sources to ensure balanced and representative coverage.

Validation and test sets were constructed via a multi-step grade- and stage-based procedure, detailed in Appendix~\ref{sec:appendix_datasheet}. University-level questions were selected chapter-wise from multiple Romanian medical schools, as outlined in the Appendix.

Originally designed with 1,400 questions each, the validation and test sets were slightly reduced following a final round of manual deduplication. Removed duplicates were reassigned to the training set to maintain evaluation integrity.

\section{Experiments}
We conducted comparisons and benchmarks across several dimensions, including zero-shot vs. few-shot settings, group-choice heuristics, and combined vs. individual predictions. To ensure reproducibility, all experiments were run with temperature set to zero. These experiments were carried out using local hardware, a Google Colab Pro subscription, and various API/runtime services, with a total cost of \$48.73. Although we do not have an exact runtime estimate, the work was completed over 2–3 months of intermittent activity.

\subsection{Benchmarking on RoBiologyDataChoiceQA}

Acknowledging good benchmarking practices explored by \citeauthor{liang2023holistic}, \citeyear{liang2023holistic}, we evaluate multiple LLMs on the test split of the RoBiologyDataChoiceQA dataset and report their accuracies in Table \ref{tab:benchmarking-models}. The selected models include those offering accessible API usage as well as competitive open-source Romanian models. Details regarding the prompts used can be found in the Appendix (\ref{sec:appendix_prompts}).

Despite the dataset being in Romanian, the Romanian-trained models (\textit{Rogemma2, Rollama3-8B-Instruct-Imat, and Romistral-7B-Instruct}) did not show a significant advantage over multilingual or primarily English-trained models. Given their explicit training on Romanian \cite{masala2024vorbecstiromanecsterecipetrain}, we expected them to perform better due to their stronger grasp of Romanian syntax and semantics. However, the observed improvements were marginal, suggesting that language understanding alone is not enough to solve this task. Instead, performance appears to be primarily constrained by the models’ ability to reason about biological concepts and apply domain knowledge rather than by linguistic factors.

Studies \cite{Nguyen2025, Gao2024} have shown that running the same models from different providers could yield slightly different accuracies in some contexts. This was not our case, since doing this resulted in nearly identical accuracies, with variations of at most 0.04. Therefore, we do not specify the source for each model. We conduct evaluations both locally and via external providers.

\begin{table}[ht!]
\centering
\small
\adjustbox{max width=\columnwidth}{
\begin{tabular}{l c c c}
\hline
\textbf{Model} & \textbf{Single Acc.} & \textbf{Group Acc.} & \textbf{Multi Acc.} \\
\hline
gemini-2.0-flash  & \textbf{0.733} & 0.524 & \textbf{0.585} \\
gemini-2.0-flash-exp & 0.719 & \textbf{0.537} & 0.539 \\
qwen-max-2025-01-25 & 0.699 & 0.472 & 0.573 \\
llama-3.1-405B-Instruct-Turbo & 0.685 & 0.426 & 0.464 \\
gemini-1.5-flash & 0.668 & 0.419 & 0.406 \\
DeepSeek-V3 & 0.665 & 0.453 & 0.474 \\
llama-3.3-70B-Instruct-Turbo & 0.629 & 0.413 & 0.378 \\
rogemma2-9b-instruct (Q8) & 0.543 & 0.298 & 0.198 \\
gemma2-9b-it & 0.529 & 0.346 & 0.226 \\
llama3-8b-instruct & 0.405 & 0.250 & 0.093 \\
phi-3.5-mini-instruct (F32) & 0.379 & 0.208 & 0.080 \\
eurollm-9b-instruct (F16) & 0.384 & 0.220 & 0.102 \\
rollama3-8b-instruct-imat (FP16) & 0.371 & 0.235 & 0.102 \\
romistral-7b-instruct (Q8) & 0.371 & 0.252 & 0.077 \\
mistral-7b-instruct-v0.1 (Q8) & 0.221 & 0.199 & 0.046 \\
\hline
Baseline & 0.245 & 0.200 & 0.032 \\
\hline
\end{tabular}
}
\caption{Accuracies of models benchmarked on zero shot.}
\label{tab:benchmarking-models}
\end{table}

\begin{figure}[h!]
  \centering
  \includegraphics[width=\columnwidth]{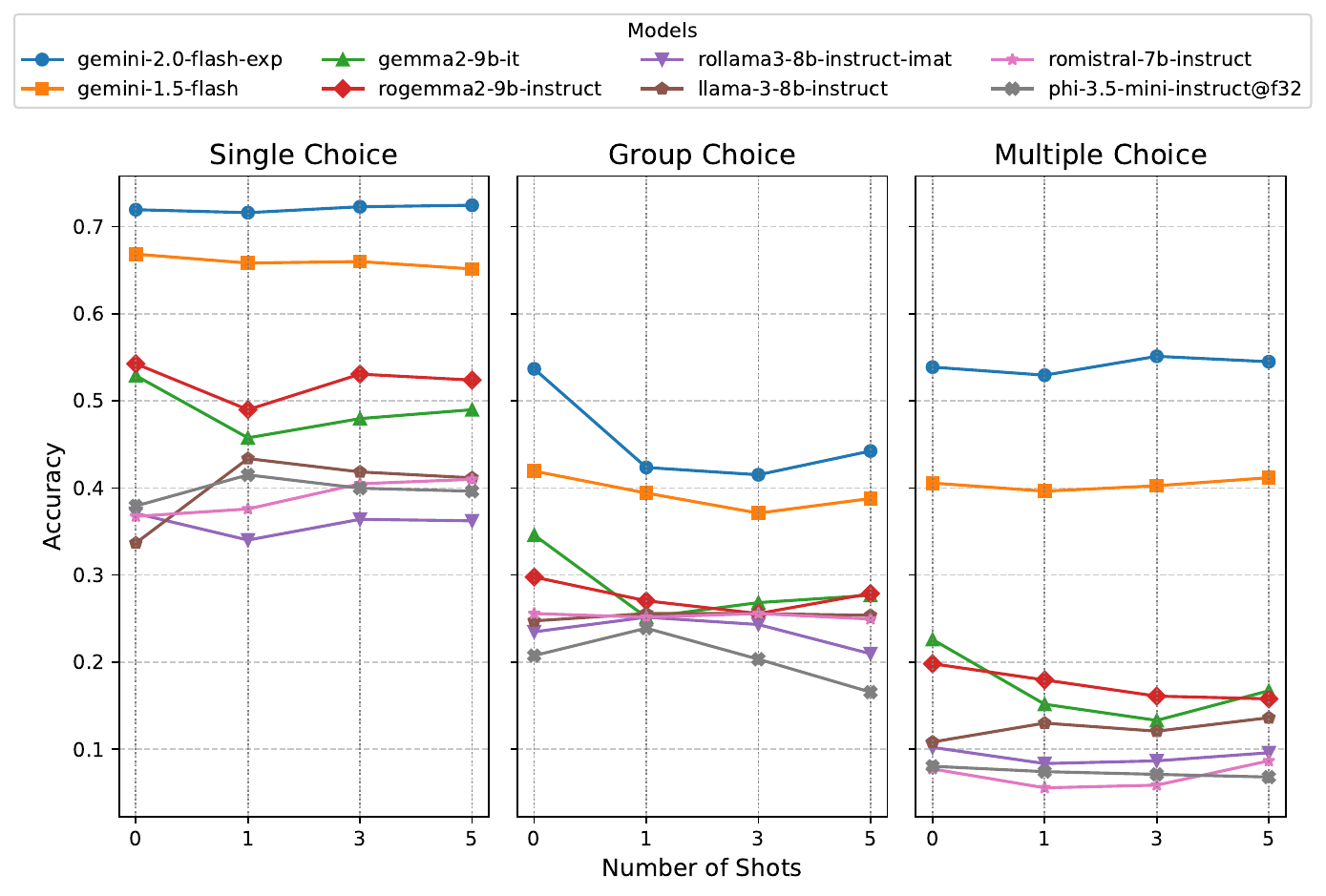}
  \caption{Accuracies of some models over few shot prompting.}
  \label{fig:num_of_shots}
\end{figure}

Running the models with a few-shot approach did not yield substantial improvements (phenomenon also found in \citeauthor{hendrycks2021measuring}, \citeyear{hendrycks2021measuring} and \citeauthor{kojima2023largelanguagemodelszeroshot}, \citeyear{kojima2023largelanguagemodelszeroshot}); in fact, some models performed worse, as shown in Figure \ref{fig:num_of_shots}. Notably, certain LLMs exhibited a tendency to overfixate on specific letters after being presented with examples—interestingly, not necessarily the ones included in the prompt. The few-shot examples were provided to the LLMs within the system prompt, as described in Appendix \ref{sec:appendix_prompts}.

\subsection{Benchmarking by source type}

\begin{table}[ht!]
\centering
\small
\adjustbox{max width=\columnwidth}{
\begin{tabular}{l cc cc}
\hline
\textbf{Multiple} & \multicolumn{2}{c}{\textbf{Single Acc.}} & \multicolumn{2}{c}{\textbf{Multiple Acc.}} \\
\hline
 & \textbf{Olympiad} & \textbf{UMF Brasov} & \textbf{UMF Timisoara} & \textbf{UMF Cluj} \\
\hline

gemini-2.0-flash-exp & 0.704 & 0.824 & 0.615 & 0.415 \\
qwen-max-2025-01-25 & 0.679 & 0.838 & 0.655 & 0.439 \\
llama-3.1-405B-Instruct-Turbo & 0.665 & 0.824 & 0.565 & 0.301 \\
gemini-1.5-flash & 0.658 & 0.743 & 0.485 & 0.276 \\
DeepSeek-V3 & 0.650 & 0.770 & 0.540 & 0.366 \\
llama-3.3-70B-Instruct-Turbo & 0.611 & 0.757 & 0.445 & 0.268 \\
rogemma2-9b-instruct (Q8) & 0.531 & 0.622 & 0.230 & 0.146 \\
gemma2-9b-it & 0.502 & 0.716 & 0.255 & 0.179 \\
llama3-8b-instruct & \cellcolor{cyan!20}0.409 & 0.378 & 0.130 & 0.033 \\
eurollm-9b-instruct (F16) & \cellcolor{cyan!20}0.393 & 0.270 & 0.110 & 0.073 \\
phi-3.5-mini-instruct (F32) & \cellcolor{cyan!20}0.387 & 0.324 & 0.085 & 0.073 \\
romistral-7b-instruct (Q8) & \cellcolor{cyan!20}0.374 & 0.324 & 0.085 & 0.065 \\
rollama3-8b-instruct-imat (FP16) & \cellcolor{cyan!20}0.372 & 0.365 & 0.120 & 0.073 \\
mistral-7b-instruct-v0.1 (Q8) & 0.210 & 0.297 & 0.055 & 0.033 \\
\hline
Baseline & 0.250 & 0.200 & 0.032 & 0.032 \\
\hline
\end{tabular}
}
\caption{Accuracies of models, separated by source.}
\label{tab:benchmarking-models-by-stage}
\end{table}

We compare model performance on Olympiad data versus university admission data. As shown in Figure \ref{tab:benchmarking-models-by-stage}, models tend to perform better on university-level questions with a single correct answer, suggesting they are more accustomed to medical admission data than to biology Olympiad questions. Alternatively, this may indicate that olympiad questions are potentially more challenging, requiring deeper knowledge and reasoning skills.

In Figure \ref{tab:benchmarking-models-by-stage}, we highlight instances where Olympiad scores surpass university admission scores. Even in these cases, the difference is generally small. However, when university admission scores are higher, the margin tends to be larger.

Comparing the difficulty levels of the three universities, we observe that the UMF Brașov exam appears to be the easiest, as it consists solely of single-answer questions. In contrast, the UMF Timișoara and UMF Cluj exams contain multiple-answer questions, making them more challenging and not directly comparable to UMF Brașov. Additionally, UMF Cluj’s exam seems to be the most difficult, as all models achieve higher scores on UMF Timișoara’s admission questions. This aligns with the common perception that among the three universities analyzed, UMF Cluj has the most difficult admission exam, followed by UMF Timișoara, while UMF Brașov is considered the easiest.

\subsection{Finetuning Gemini 1.5 Flash}

Google AI Studio allows fine-tuning of the \textbf{Gemini 1.5 Flash} model with custom data by providing a CSV file where one column serves as the input and another as the model's output. Using the training split of the RoBiologyDataChoiceQA dataset, we set the input as the benchmarking prompt, replacing \%question-text\% with the formatted question entry. The output corresponds to the correct answer field without additional formatting.

\begin{figure}[ht!]
\centering
\includegraphics[width=\columnwidth]{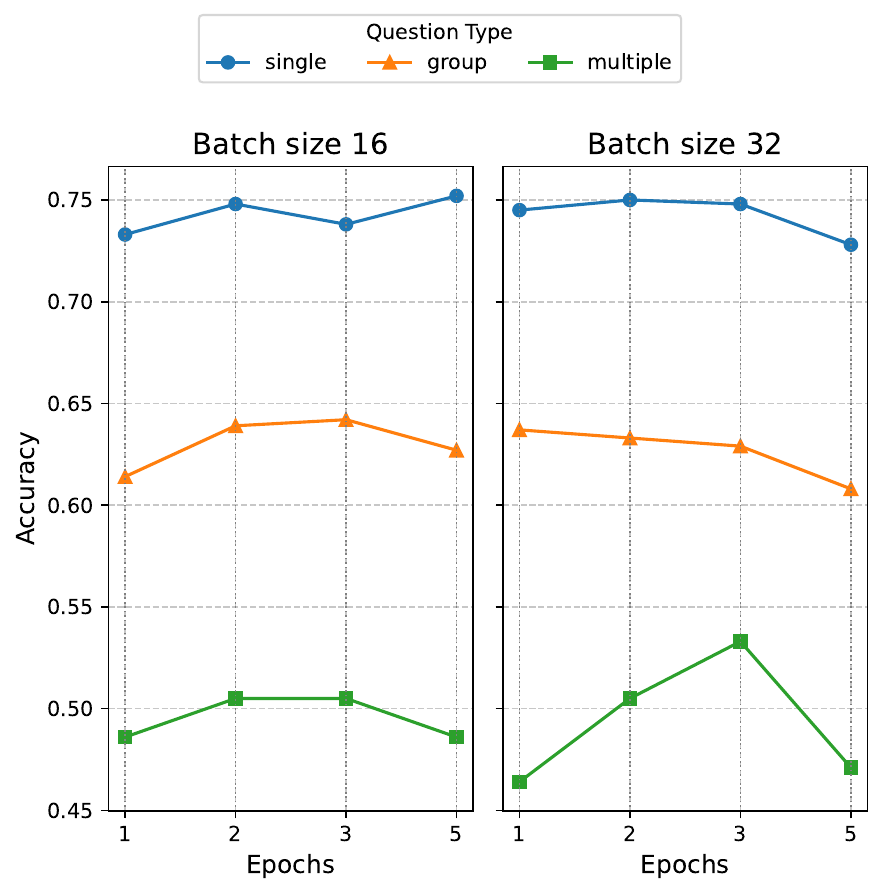}
\caption{Accuracies of fine-tuned versions of Gemini 1.5 Flash.}
\label{fig:finetune}
\end{figure}

Once training is complete, we evaluate the fine-tuned model on the test split. We train multiple versions with different parameter settings (e.g., number of epochs, batch size) as detailed in Figure \ref{fig:finetune}. Our fine-tuned models achieve new state-of-the-art accuracies, as shown in Table \ref{tab:finetune}.

\begin{table}[ht!]
\centering
\adjustbox{max width=\columnwidth}{
\begin{tabular}{lccc}
\hline
\textbf{Model} & \textbf{Single Accuracy} & \textbf{Group Accuracy} & \textbf{Multiple Accuracy} \\
\hline
gemini-2.0-flash  & 0.733 & 0.524 & \textbf{0.585} \\
\hline
tuned\_batch16\_epochs5 & \textbf{0.752} & 0.627 & 0.486 \\
tuned\_batch16\_epochs3 & 0.738 & \textbf{0.642} & 0.505 \\
tuned\_batch16\_epochs1 & 0.733 & 0.614 & 0.486 \\
tuned\_batch32\_epochs5 & 0.728 & 0.608 & 0.471 \\
tuned\_batch32\_epochs3 & 0.748 & 0.629 & 0.533 \\
tuned\_batch32\_epochs2 & 0.750 & 0.633 & 0.505 \\
tuned\_batch32\_epochs1 & 0.745 & 0.637 & 0.464 \\
tuned\_batch16\_epochs2 & 0.748 & 0.639 & 0.505 \\
tuned\_batch64\_epochs3 & 0.733 & 0.612 & 0.517 \\
\hline
gemini-1.5-flash & 0.668 & 0.419 & 0.406 \\
\hline
\end{tabular}
}
\caption{Accuracies of fine-tuned Gemini 1.5 Flash models}
\label{tab:finetune}
\end{table}

\subsection{Finetuning Gemma 2 9B Instruct}

\begin{figure}[ht!]
\centering
\includegraphics[width=\columnwidth]{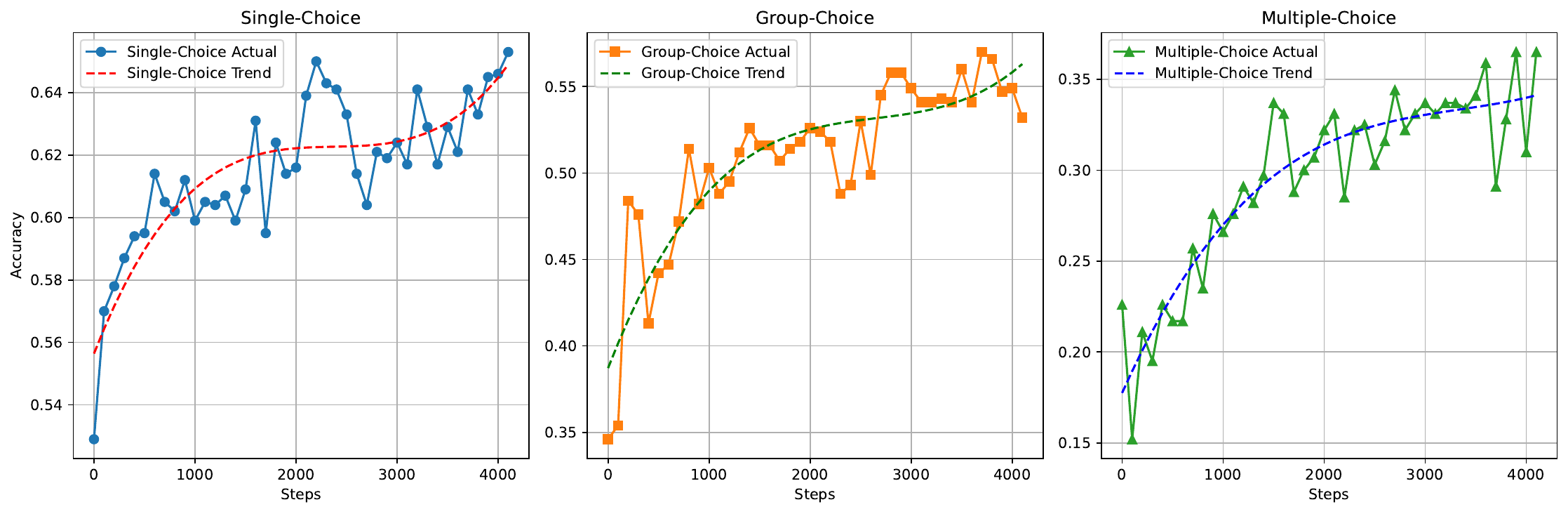}
\caption{Performance of Gemma 2 9B Instruct on the test split over fine-tuning training steps.}
\label{fig:gemma-finetune}
\end{figure}

After successfully improving Gemini’s performance through fine-tuning, we extend this approach to a smaller model, Gemma 2 9B Instruct, and observe similar accuracy gains, as shown in Figure \ref{fig:gemma-finetune}.

For fine-tuning, we employ the LoRA technique \cite{hu2021loralowrankadaptationlarge} via the Unsloth framework \cite{unsloth}, training the model for approximately four epochs, with 1,000 steps per epoch. Accuracy is evaluated at intervals of 100 steps. While we halted training at four epochs, the observed trend suggests that further improvements may still be possible, particularly for single-choice and group-choice questions.

\begin{table}[ht!]
\centering
\adjustbox{max width=\columnwidth}{
\begin{tabular}{lccc}
\hline
 & \textbf{Single Acc.} & \textbf{Group Acc.} & \textbf{Multiple Acc.} \\
\hline
gemma2-9b-it & 0.529 & 0.346 & 0.226 \\
\hline
finetune step 3700 & 0.641 & 0.\textbf{570} & 0.291 \\
finetune step 3900 & 0.645 & 0.547 & \textbf{0.365} \\
finetune step 4100 & \textbf{0.653} & 0.532 & \textbf{0.365} \\
\hline
max increase & 0.124 & 0.186 & 0.139 \\
\hline

\hline
\end{tabular}
}
\caption{Best accuracies of the model during fine-tuning.}
\label{tab:gemma-finetune}
\end{table}

Table \ref{tab:gemma-finetune} reports the highest accuracies obtained during fine-tuning. Compared to the initial model, Gemma 2 9B Instruct achieves improvements of over 12 percentage points. The fine-tuned model attains performance comparable to larger models, significantly narrowing the gap with Gemini 1.5 Flash on single-choice and multiple-choice questions (falling behind by only 1.5 and 4.1 percentage points, respectively). For group-choice questions, it outperforms all models from the initial benchmark, surpassing the previous state-of-the-art by 3.3 percentage points.

\subsection{Treating group choice questions as multiple choice}

Inspired by \citeauthor{balepur-etal-2024-artifacts}, \citeyear{balepur-etal-2024-artifacts}, we hypothesized that LLMs might struggle to correctly apply the grouping rules, particularly in cases where the multiple-choice accuracy was higher. To test this, we reformulated the questions into a multiple-choice format, ran them as if they were multiple-choice questions, and then manually mapped the groupings to their respective answers.

For cases where the model produces invalid combinations that cannot be mapped to a valid answer, we select the first letter (essentially randomizing the answer). This results in a new accuracy, which sometimes exceeds the original.

To further improve this accuracy, we implemented heuristics instead of relying on the random approach for invalid groups. For example, the combination (1, 2) is mapped to (1, 2, 3); (1) or (3) is mapped to (1, 3); (2, 3, 4) is mapped to (1, 2, 3, 4), and so on. For most models, the use of heuristics yields better results than the random selection, as shown in Table \ref{tab:group-as-multiple}.

\begin{table}[ht!]
\centering
\adjustbox{max width=\columnwidth}{
\begin{tabular}{lccc}
\hline
\textbf{Model} & \textbf{Group} & \textbf{Group As Multiple} & \textbf{With Heuristics} \\
\hline
gemini-2.0-flash-exp & 0.537 & 0.449 & 0.499 \\
DeepSeek-V3 & 0.453 & 0.388 & 0.423 \\
llama-3.1-405B-Instruct-Turbo & 0.426 & \cellcolor{cyan!20}0.453 & 0.484\\
gemini-1.5-flash & 0.419 &\cellcolor{cyan!20} 0.447 & 0.480 \\
gemma2-9b-it & 0.346 & 0.300 & 0.314 \\
rogemma2-9b-instruct (Q8) & 0.298 & 0.258 & 0.275 \\
llama3-8b-8192 & 0.252 & 0.235 & 0.245 \\
rollama3-8b-instruct-imat (FP16) & 0.235 & \cellcolor{cyan!20}0.241 & 0.256 \\
phi-3.5-mini-instruct (F32) & 0.208 & \cellcolor{cyan!20}0.231 & 0.247 \\
\hline
\end{tabular}
}
\caption{The accuracies obtained on group choice questions with all strategies. Highlighting signifies a better score with the group-as-multiple approach compared to the initial strategy.}
\label{tab:group-as-multiple}
\end{table}

\subsection{Accuracy by Grade}

We also compare the accuracies obtained on questions, grouped by the corresponding grade level.

\begin{figure}[h!]
  \centering
  \includegraphics[width=\columnwidth]{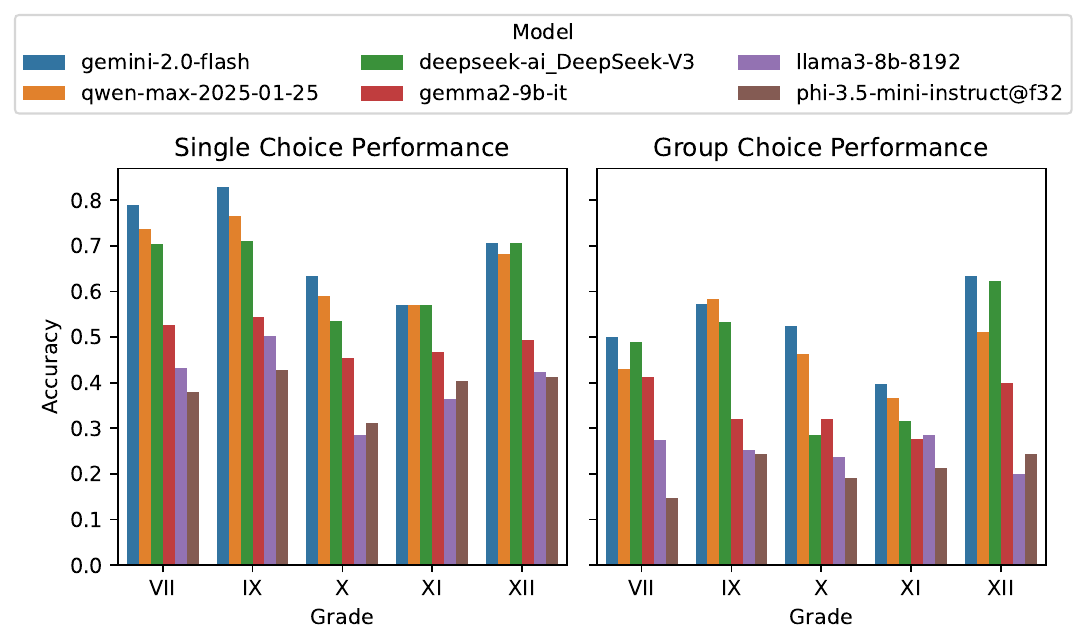}
  \caption{Accuracies of models, grouped by competition grade}
  \label{fig:accuracy-by-grade}
\end{figure}

As shown in Figure \ref{fig:accuracy-by-grade}, models achieve the lowest scores on grades X and XI, while performing better on grades IX and XII. Performance on grade VII falls between these extremes.

Examining the curricula for these grade levels, we observe a correlation between subject focus and model accuracy. Grades IX and XII emphasize molecular biology and interactions between biological systems, while grades X and XI focus on the physiology and functions of biological systems (see examples in Figure~\ref{fig:examples-by-grades}). Grade VII provides a broad introduction, covering aspects of all these topics while also including basic principles of hygiene and health.

These results suggest that models perform better on topics related to molecular biology and genetics compared to those centered on the physiology of biological systems.

\subsection{Accuracy by Stage}

\begin{figure}[h!]
  \centering
  \includegraphics[width=\columnwidth]{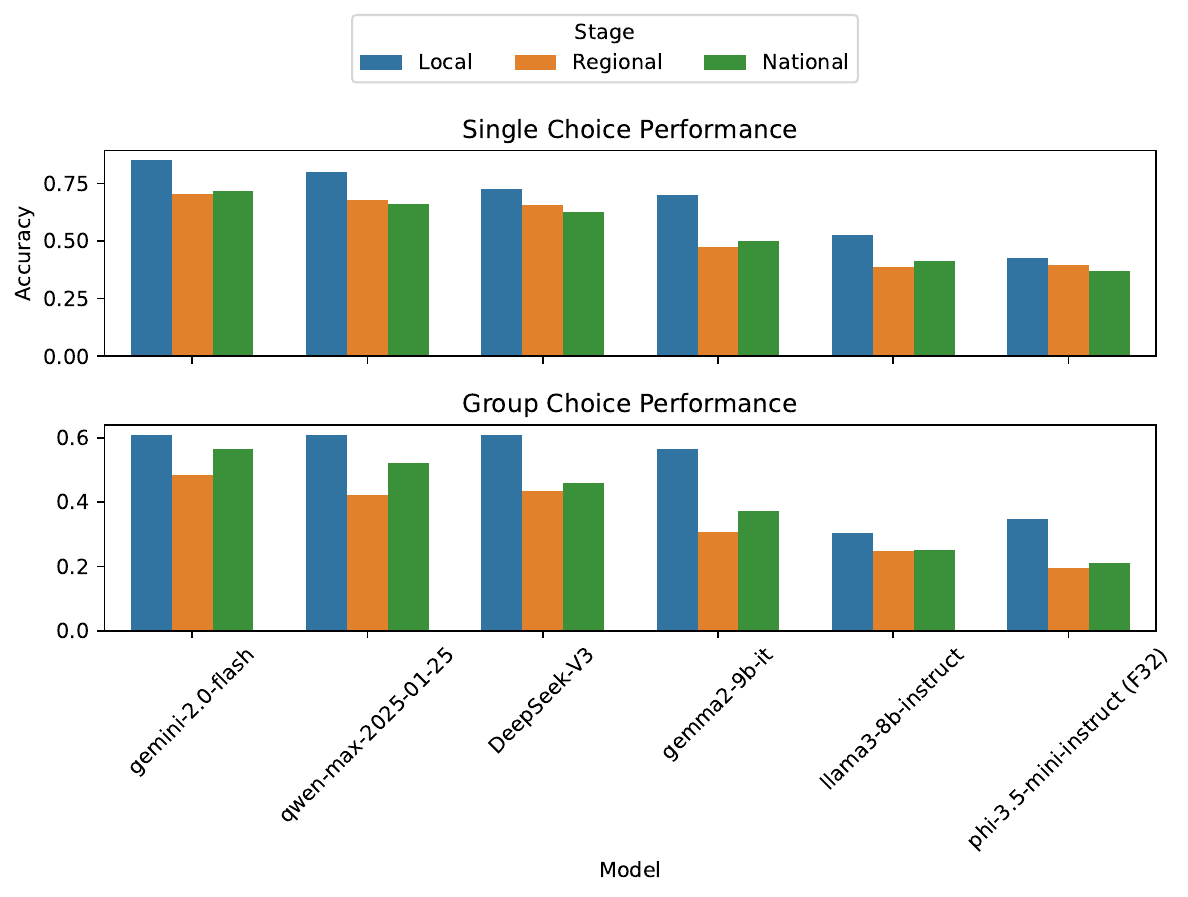}
  \caption{Accuracies of models on different competition stages.}
  \label{fig:accuracy-by-stage}
\end{figure}

We compare the accuracies obtained on questions from the test split, grouped by the competition stage in which they were presented (local, regional, or national), and report the results in Figure \ref{fig:accuracy-by-stage}.

For both single-answer and group-choice questions, models achieve the highest scores on the local stage, confirming that it is indeed the easiest of the three. For single-choice questions, the accuracy remains similar between the regional and national stages, suggesting comparable difficulty levels. However, for group-choice questions, models unexpectedly perform better on the national stage than on the regional stage, despite the expectation that the national stage should be more challenging.

\subsection{Model Ensemble}

Building on the LLM-Synergy framework proposed by \citeauthor{Yang2023.12.21.23300380} (\citeyear{Yang2023.12.21.23300380}), who used Majority Weighted Voting to aggregate outputs from multiple LLMs for biomedical QA, we implemented a simplified ensemble learning strategy to enhance model performance on our dataset. Specifically, we created three groups of models with comparable individual accuracies: (1) top-performing models, (2) mid-range models, and (3) models fine-tuned on Romanian. Each group included three models, allowing us to use unweighted Majority Voting, as weighting would not affect the outcome. All experiments were conducted under zero-shot settings and computed separately for single, group, and multiple-choice questions.

Table \ref{tab:combined_pred1}, \ref{tab:combined_pred2}, and \ref{tab:combined_pred3} present the results of these ensemble experiments.

Although not by a significant difference, the Majority Voting surpassed the individual performances on group-choice questions in all of the chosen model subsets.

\begin{table}[ht!]
\centering
\adjustbox{max width=\columnwidth}{
\begin{tabular}{lccc}
\hline
\textbf{Model} & \textbf{Single} & \textbf{Group} & \textbf{Multiple} \\
\hline
gemini-2.0-flash & \textbf{0.733} & 0.524 & \textbf{0.585} \\
qwen-max-2025-01-25 & 0.699 & 0.472 & 0.573 \\
llama-3.1-405B-Instruct-Turbo & 0.685 & 0.426 & 0.464 \\
All of the above combined & 0.719 & \textbf{0.534} & 0.560 \\

\hline
\end{tabular}
}
\caption{The accuracy of Majority Voting compared to the individual accuracies.}
\label{tab:combined_pred1}
\end{table}

\begin{table}[ht!]
\centering
\adjustbox{max width=\columnwidth}{
\begin{tabular}{lccc}
\hline
\textbf{Model} & \textbf{Single} & \textbf{Group} & \textbf{Multiple} \\
\hline
DeepSeek-V3 & 0.665 & 0.453 & \textbf{0.474} \\
gemini-1.5-flash & 0.668 & 0.419 & 0.406 \\
llama-3.1-405B-Instruct-Turbo & 0.685 & 0.426 & 0.464 \\
All of the above combined & \textbf{0.707} & \textbf{0.457} & 0.439 \\

\hline
\end{tabular}
}
\caption{The accuracy of Majority Voting compared to the individual accuracies.}
\label{tab:combined_pred2}
\end{table}

\begin{table}[ht!]
\centering
\adjustbox{max width=\columnwidth}{
\begin{tabular}{lccc}
\hline
\textbf{Model} & \textbf{Single} & \textbf{Group} & \textbf{Multiple} \\
\hline
eurollm-9b-instruct (F16) & \textbf{0.384} & 0.220 & \textbf{0.102} \\
rollama3-8b-instruct-imat (FP16) & 0.371 & 0.235 & \textbf{0.102} \\
romistral-7b-instruct (Q8) & 0.371 & 0.252 & 0.077 \\
All of the above combined & 0.372 & \textbf{0.266} & \textbf{0.102} \\

\hline
\end{tabular}
}
\caption{The accuracy of Majority Voting compared to the individual accuracies.}
\label{tab:combined_pred3}
\end{table}

\subsection{Error Analysis}

To explore common failure patterns in LLMs, we analyzed 75 questions that were incorrectly answered by all benchmarked models (24 single-choice, 8 group-choice, 43 multiple-choice).

We presented these questions to a medical student and observed that when asked to respond quickly, their responses often resembled those of the models. However, when given more time, the student changed several responses. This indicates a potential need for models to also ponder their responses, which we did not sufficiently investigate (using techniques like multi-turn prompting or thinking tokens).

Beyond this, we observed that models often rely on superficial associative reasoning. For instance, when prompted with “Hiperglicemia poate determina o:” (“Hyperglycemia can determine a:”), models alternated between “hyposecretion of insulin” and “hypersecretion of glucagon,” whereas the correct answer was “hyposecretion of glucocorticoids.” We hypothesize this results from a bias toward more frequently co-occurring hormone-glucose relations in public corpora, and a lack of exposure to nuanced clinical cases.

Models also struggle with traps involving lexical similarity or subtle qualifiers. All failed a question by confusing “bronhii” (bronchi) with “bronhiole” (bronchioles). In another, most selected “gravitational pull for veins located below the heart level” as promoting venous return, an incorrect answer due to the phrasing “below” instead of “above”. These patterns suggest a lack of deeper contextual reasoning.
\section{Conclusion}

This study introduced RoBiologyDataChoiceQA, a novel Romanian-language dataset designed to evaluate biology comprehension in large language models (LLMs). Sourced from the Romanian Biology Olympiad and medical school entrance exams, it provides a diverse and challenging benchmark for assessing domain-specific reasoning in a low-resource language.

Our benchmarking experiments revealed significant variations in model performance, highlighting both strengths and limitations of LLMs in specialized tasks. While some models performed well on structured, single-answer university questions, their ability to handle grouped-choice and reasoning tasks remained inconsistent. Fine-tuning Gemini 1.5 Flash and Gemma 2 9B Instruct improved accuracy in certain cases, demonstrating that targeted adaptation can be effective.

Beyond model evaluation, our study offers insights into the impact of prompt engineering, fine-tuning strategies, and dataset characteristics on LLM performance. These findings contribute to the broader effort of advancing NLP applications in non-English languages and scientific domains.

Future research should focus on expanding the dataset with fine-grained subdomain annotations, improving OCR processing, experimenting with other fine-tuning strategies and model architectures, and addressing dataset biases by comparing performance across question sources. Enhancing answer verification through expert validation will also be essential for benchmark reliability.
\section{Limitations}

While our study provides valuable insights into LLM performance on Romanian-language biology questions, several limitations should be considered when interpreting the results.

\begin{itemize}
    \item \textbf{Limited computational resources} – Most experiments were conducted using a single NVIDIA RTX 3070 GPU (8 GB VRAM) paired with 32 GB of system RAM, along with external API and runtime providers. This constrained our ability to perform large-scale experimentation, including multiple training runs, broader hyperparameter sweeps, and evaluation of larger models.

    \item \textbf{Lack of fine-grained tagging} – The dataset does not include detailed annotations distinguishing specific biological subdomains (e.g., genetics, physiology, ecology). This limits the ability to analyze model performance at a more granular level and identify knowledge gaps in specialized areas.
    
    \item \textbf{Potential inaccuracies in answer keys} – Although we rely on authoritative sources, occasional ambiguities or errors in the provided answer keys may affect benchmarking accuracy. While we performed additional verification, some uncertainties remain.
    
    \item \textbf{Challenges with OCR-extracted data} – The dataset includes content extracted from scanned PDFs, particularly for university admission exams. Despite preprocessing and manual validation, some errors introduced by OCR remain, potentially affecting model training and evaluation.
    
    \item \textbf{Limited scope of fine-tuning experiments} While we observed improvements when fine-tuning Gemini 1.5 Flash and Gemma 2 9B Instruct, additional experiments with different architectures and training strategies could yield further insights. Exploring other Romanian-adapted models could provide a broader perspective.
    
    \item \textbf{Domain-specific biases in LLMs} – Our results suggest that models perform better on university admission questions than on Olympiad questions, likely due to differences in training data exposure. Investigating whether this bias stems from pretraining corpora, difficulty of questions, or inherent reasoning limitations could further refine model evaluation.

    \item \textbf{Language vs. Domain Effects:} We do not perform a cross-lingual evaluation (e.g., testing models on an English version of the dataset or on other Romanian-language datasets) to isolate the impact of language from domain complexity. As such, we cannot fully disentangle whether observed model weaknesses stem primarily from Romanian language handling or from the specialized nature of biology. We leave this analysis to future work.

    \item \textbf{Potential Data Leakage:} We do not explicitly verify whether the dataset’s questions appear in the training data of the evaluated language models, particularly open-weight models. Due to the lack of transparency around training corpora and the impracticality of exhaustively checking large-scale pretraining data, this remains a potential source of data leakage. While API-based models’ training data are even less accessible, we acknowledge that possible overlap could bias performance results. We consider this an important caveat and encourage future work to investigate this aspect more thoroughly.

\end{itemize}

\section{Ethical Statement}
To promote transparency and responsible use, we release the dataset under the \textit{Creative Commons Attribution-NonCommercial 4.0 International (CC BY-NC 4.0)} license. This license allows for non-commercial use, sharing, and adaptation with proper attribution.

No personally identifiable or sensitive information is included in the dataset. We encourage ethical research practices and responsible AI development when using our dataset. However, a potential risk is that it could inadvertently encourage the use of LLMs in biology exams for cheating, rather than for legitimate educational or research purposes. We urge users to adopt responsible policies to prevent misuse in academic settings.

\bibliography{anthology,custom}
\bibliographystyle{acl_natbib}

\appendix

\section{Datasheet}
\label{sec:appendix_datasheet}
\subsection{Motivation for Dataset Creation}
\bigskip 
\textbf{Why was the dataset created?}

The dataset was developed to assess and enhance the performance of large language models (LLMs) on domain-specific tasks, specifically Romanian biology tests. It offers choice-based questions to evaluate LLM accuracy and can also be used for fine-tuning LLMs to understand specialized Romanian biology terminology.
\bigskip 

\textbf{What (other) tasks could the dataset be used for?}

One potential application of this dataset is its use as training data for models designed to generate multiple-choice questions. Additionally, the dataset could be utilized for automatically assessing question difficulty.
\bigskip

\subsection{Dataset Composition}
\bigskip

\textbf{What are the instances?}

The instances consist of (single, group, or multiple) choice questions sourced from Romanian biology olympiads and college admission exam books. Each question is paired with its correct answer(s), extracted from the corresponding answer keys. Additional identifying information is also appended to each instance, as detailed in the following paragraphs.
\bigskip

\textbf{Are relationships between instances made explicit in the data?}

Yes, relationships between instances are explicitly marked. Using question identification metadata, instances can be grouped by attributes such as source, year, grade, and stage. When identical questions with identical answer options appear across different tests or problem sets, they are assigned a shared \textit{dupe\_id}.

Duplicates are retained rather than removed for several reasons:
\begin{itemize}
    \item To analyze patterns of data repetition (e.g., identifying sources of inspiration between tests).
    \item To avoid arbitrarily deciding which instance to delete, leaving duplicate removal to the user's discretion.
\end{itemize}
All known duplicates are included exclusively in the training split.
\bigskip

\textbf{How many instances of each type are there?}

The dataset contains a total of 14,109 extracted questions:
\begin{itemize}
    \item Single choice: 6,021
    \item Group choice: 3,918
    \item Multiple choice: 4,170
\end{itemize}

Of these, 8,021 questions are sourced from biology olympiads, while 6,088 come from college admission books. The tests span multiple years (2004-2024), although they are not uniformly distributed.
\bigskip

\textbf{What data does each instance consist of?}

We will explain each field:
\begin{itemize}
    \item \textbf{question\_number} = an integer stored as string; for olympiads it takes values from 1 to 80. Most tests tend to have at most 60, but the very old ones (2004) do not quite respect the format. As for college admissions, those take values from 1 to 800 (not uniformly, there are tests/chapters with random number of questions, no general rule).
    \item \textbf{question} = the question text
    \item \textbf{type} - can be one of the following:
        \begin{itemize}
            \item \textit{single-choice}: indicating the question has exactly one correct answer.
            \item \textit{group-choice}: indicating that the answer is a single letter, which corresponds to a combination of options being true together:
               \begin{quote}
                   \textbf{A} - if ONLY the options numbered by 1, 2 and 3 are correct\\
                   \textbf{B} - if ONLY the options numbered by 1 and 3 are correct\\
                   \textbf{C} - if ONLY the options numbered by 2 and 4 are correct\\
                   \textbf{D} - if ONLY the option numbered by 4 is correct\\
                   \textbf{E} - if ALL of the numbered options are correct\\
               \end{quote}
   The group choice is the only type that has options identified by numbers, while the others have them identified by letters.
            \item \textit{multiple-choice}: indicating that the answer is represented by any alphabetically ordered combination of the given options. Even though it is multiple, the answer CAN STILL be a single letter)
          \end{itemize}
    \item \textbf{options} = a list of texts (usually statements or list of items) that in combination with the question text can be considered true or false. Olympiad tests have 4 options, while college admission tests have 5.
    \item \textbf{grade} = where the test/problem set was extracted from; it takes 6 values: \textit{facultate} (college), \textit{XII}, \textit{XI}, \textit{X}, \textit{IX} (highschool), \textit{VII} (middle school).
    \item \textbf{stage} = for college it is fixed on \textit{admitere} (admission). For olympiad it represents the chain of theoretical importance and difficulty: \textit{locala -> judeteana -> nationala} (local -> regional -> national).
    \item \textbf{year} = the year (as a string) in which the problem set/test was used in a competition
    \item \textbf{right\_answer} = a letter for single-choice and group-choice (check the explanations above) and multiple (non-repeating) letters concatenated in a string with no other characters, in alphabetical order for multiple-choice.
    \item \textbf{source} = \textit{olimpiada} (Olympiad of Biology in Romania) or, in the case of college, the university it was taken from (currently 3 possible values: \textit{UMF Cluj}, \textit{UMF Brașov}, \textit{UMF Timișoara})
    \item \textbf{id\_in\_source} = a string that has the purpose of further recognising the question within the problem set it was given, in case of ambiguity. Ensures uniqueness when combined with the other fields recommended for identifying the questions. Keep in mind that it contains spaces.
    \item \textbf{dupe\_id} = a UUID that uniquely identifies a group of duplicated questions. The group may contain 2 or more instances. The instance is considered a duplicate if and only if both the question and options are the same (not necessarily in the same order for options). Two texts are considered the same if they are identical/use synonyms for common words/are obviously rephrased versions of each other. If a text adds extra words but besides that it is identical with another text, it is \textit{not} marked as a duplicate.
\end{itemize}

For uniquely identifying a question/instance we recommend the following combination of fields:

\[
\left\{
\begin{array}{l}
    \texttt{item['year']}, \\
    \texttt{item['source']}, \\
    \texttt{item['id\_in\_source']}, \\
    \texttt{item['grade']}, \\
    \texttt{item['stage']}, \\
    \texttt{item['question\_number']}
\end{array}
\right\}
\]
\bigskip

\textbf{Is everything included or does the data rely on external resources?}

Everything is included.
\bigskip

\textbf{Are there recommended data splits or evaluation measures?}

The data is currently split into three: train, valid, test. We attempted a uniform distribution of the data, based on both quantity and quality of the data.

Both the \textit{test} and \textit{valid} splits were sampled via the recipe explained below.

First we do a grade-based separation:
\begin{itemize}
    \item Grade XII: 175 questions  \\
  - 75 national level  \\
  - 100 state level  
    \item Grade XI: 175 questions \\
  - 75 national level \\
  - 100 state level 
    \item Grade X: 200 questions \\
  - 55 national level \\
  - 125 state level \\
  - 20 local level  
    \item Grade IX: 250 questions \\
  - 115 national level \\
  - 115 state level \\
  - 20 local level  
    \item Grade VII: 200 questions \\
  - 85 national level \\
  - 85 state level \\
  - 30 local level  
    \item University Level (\textit{Facultate}): 400 questions (detailed division below)
\end{itemize}

1. \textit{UMF Timișoara}: 200 questions \\
   - 11 chapters total, 18 questions per chapter, except for the \textit{Nervous System}, which has 20 questions due to higher coverage.  
   
\smallskip

2. \textit{UMF Brașov}: 75 questions \\
   - Derived from 15 questions from each synthesis test.  

\smallskip

3. \textit{UMF Cluj}: 125 questions \\
   - \textit{Physiology} (for medical assistant students): 8 questions (1 question per chapter for 5 chapters, plus 3 random questions) \\
   - \textit{Anatomy} (for medical assistant students): 8 questions (same structure as \textit{Physiology}) \\
   - \textit{Physiology} (for medical students): 55 questions (4 questions from each of the first 13 chapters, plus 3 questions from Chapter 14) \\
   - \textit{Anatomy} (for medical students): 54 questions (similar to \textit{Physiology}, but only 2 questions from Chapter 14) \\
\bigskip

\textbf{Grade-Stage Yearly Distribution}

The tables \ref{tab:year_stage_national}, \ref{tab:year_stage_regional}, \ref{tab:year_stage_local} present the yearly distribution of how many questions to select for each grade, per stage: “\textit{-}” means no data was available for that year, while “\textit{X}” means nothing was selected.

\begin{table*}
\centering
\resizebox{\textwidth}{!}{
\begin{tabular}{|c|c|c|c|c|c|c|c|c|c|c|c|c|c|c|c|c|c|c|c|c|c|c|c|c|}
\hline
 \rowcolor{gray!20} & \textbf{04} & \textbf{05} & \textbf{06} & \textbf{07} & \textbf{08} & \textbf{09} & \textbf{10} & \textbf{11} & \textbf{12} & \textbf{13} & \textbf{14} & \textbf{15} & \textbf{16} & \textbf{17} & \textbf{18} & \textbf{19} & \textbf{20} & \textbf{21} & \textbf{22} & \textbf{23} & \textbf{24} \\ \hline
\cellcolor{gray!20} \textbf{VII} & - & - & - & - & - & 5 & 5 & 7 & 8 & 8 & 12 & 15 & 15 & - & - & - & - & - & - & - & - \\ \hline
\cellcolor{gray!20} \textbf{IX} & 2 & 2 & - & - & 4 & 4 & - & 5 & 5 & 5 & 8 & 8 & 8 & - & 10 & 12 & - & - & 12 & 15 & 15 \\ \hline
\cellcolor{gray!20} \textbf{X} & - & - & - & - & - & - & - & - & - & - & 3 & 3 & 4 & - & 5 & 7 & - & - & 8 & 10 & 15 \\ \hline
\cellcolor{gray!20} \textbf{XI} & - & - & - & - & - & - & - & - & - & - & 5 & 5 & 7 & - & 8 & 8 & - & - & 12 & 15 & 15 \\ \hline
\cellcolor{gray!20} \textbf{XII} & - & - & - & - & - & - & - & - & - & - & 5 & 5 & 7 & - & 8 & 8 & - & - & 12 & 15 & 15 \\ \hline
\end{tabular}
}
\caption{Number of questions to select in test/validation data for each grade in every year from the \textbf{national} stage of the olympiad.}
\label{tab:year_stage_national}
\end{table*}

\begin{table*}
\centering
\resizebox{\textwidth}{!}{
\begin{tabular}{|c|c|c|c|c|c|c|c|c|c|c|c|c|c|c|c|c|c|c|c|c|c|c|c|c|}
\hline
 \rowcolor{gray!20} & \textbf{04} & \textbf{05} & \textbf{06} & \textbf{07} & \textbf{08} & \textbf{09} & \textbf{10} & \textbf{11} & \textbf{12} & \textbf{13} & \textbf{14} & \textbf{15} & \textbf{16} & \textbf{17} & \textbf{18} & \textbf{19} & \textbf{20} & \textbf{21} & \textbf{22} & \textbf{23} & \textbf{24} \\ \hline
\cellcolor{gray!20} \textbf{VII} & - & - & - & - & - & 5 & 5 & 7 & 8 & 12 & 13 & 15 & - & - & - & - & - & - & - & - & - \\ \hline
\cellcolor{gray!20} \textbf{IX} & 1 & 1 & - & - & 1 & 2 & 2 & 3 & 3 & 3 & 4 & 4 & 6 & 8 & 10 & 12 & 12 & - & 13 & 15 & 15 \\ \hline
\cellcolor{gray!20} \textbf{X} & - & - & - & - & - & - & - & - & - & - & 5 & 5 & 6 & 8 & 10 & 12 & 14 & - & 20 & 20 & 25 \\ \hline
\cellcolor{gray!20} \textbf{XI} & - & - & - & - & - & - & - & - & - & - & 4 & 4 & 6 & 8 & 8 & 12 & 14 & - & 14 & 15 & 15 \\ \hline
\cellcolor{gray!20} \textbf{XII} & - & - & - & - & - & - & - & - & - & - & 4 & 4 & 6 & 8 & 8 & 12 & 14 & - & 14 & 15 & 15 \\ \hline
\end{tabular}
}
\caption{Number of questions to select in test/validation data for each grade in every year from the \textbf{regional} stage of the olympiad.}
\label{tab:year_stage_regional}
\end{table*}

\begin{table*}
\centering
\resizebox{\textwidth}{!}{
\begin{tabular}{|c|c|c|c|c|c|c|c|c|c|c|c|c|c|c|c|c|c|c|c|c|c|c|c|c|}
\hline
 \rowcolor{gray!20} & \textbf{04} & \textbf{05} & \textbf{06} & \textbf{07} & \textbf{08} & \textbf{09} & \textbf{10} & \textbf{11} & \textbf{12} & \textbf{13} & \textbf{14} & \textbf{15} & \textbf{16} & \textbf{17} & \textbf{18} & \textbf{19} & \textbf{20} & \textbf{21} & \textbf{22} & \textbf{23} & \textbf{24} \\ \hline
\cellcolor{gray!20} \textbf{VII} & X & - & - & - & - & X & X & - & X & X & X & X & X & 15 & 15 & - & - & - & - & - & - \\ \hline
\cellcolor{gray!20} \textbf{IX} & X & - & - & - & - & X & - & - & X & X & X & X & X & 15 & 15 & - & - & - & - & - & - \\ \hline
\cellcolor{gray!20} \textbf{X} & X & - & - & - & - & X & - & - & X & X & X & - & X & 10 & 10 & - & - & - & - & - & - \\ \hline
\cellcolor{gray!20} \textbf{XI} & - & - & - & - & - & - & - & - & - & - & - & - & - & - & - & - & - & - & - & - & - \\ \hline
\cellcolor{gray!20} \textbf{XII} & - & - & - & - & - & - & - & - & - & - & - & - & - & - & - & - & - & - & - & - & - \\ \hline
\end{tabular}
}
\caption{Number of questions to select in test/validation data for each grade in every year from the \textbf{local} stage of the olympiad.}
\label{tab:year_stage_local}
\end{table*}

\begin{quote}
\textbf{Note:} While each split originally contained 1,400 questions (summing everything mentioned above), the validation and test splits have fewer questions than expected. Although duplicates were identified prior to splitting, an additional round of manual duplicate verification was conducted specifically for the validation and test sets. Newly identified duplicates were moved to the training split, reducing the size of the validation and test splits.
\end{quote}
\bigskip

\subsection{Data Collection Process}
\bigskip

\textbf{How was the data collected?}

\textit{Olympiad data}: Sourced from public online archives, primarily from \textit{olimpiade.ro} (\url{https://www.olimpiade.ro/}). Additional data was retrieved through separate online searches when needed.

\textit{College admission books}: Obtained from the internet. The collected data consists of 
PDFs, with some containing parsable text and others consisting of images that required additional processing.
\bigskip

\textbf{Who was involved in the data collection process?}

The PDF data was collected by our team, with guidance from medical students who provided valuable insights on where to locate the relevant materials.
\bigskip

\textbf{Over what time-frame was the data collected?}

It took roughly one month to collect the data.
\bigskip

\textbf{How was the data associated with each instance acquired?}

The data was initially collected as PDF files. To standardize the format, a Word-to-PDF converter was sometimes used. The PDFs either contained parsable text or had text embedded in images. While the quality of some images was questionable, most of the information was successfully recognized.

For PDFs with parsable text, Python libraries were used for data extraction, with occasional manual verification and refactoring. For PDFs containing images, Gemini 1.5 Flash was employed to extract the data. Random sampling was performed to verify the accuracy of the extracted data.
\bigskip

\textbf{Does the dataset contain all possible instances?}

No. Some olympiads, although we know for sure existed, were not found on the internet. Additionally, there is more data collected in PDF format that has not yet been parsed into actual instances.
\bigskip

\textbf{If the dataset is a sample, then what is the population?}

The population includes additional college admissions and olympiads from Romania that can be found and parsed. It can also contain closely related national contests that feature choice-based questions, which could be included.
\bigskip

\textbf{Is there information missing from the dataset and why?}

Questions that included images/figures were removed as this is not a multi-modal dataset (at the moment). 
\bigskip

\textbf{Are there any known errors, sources of noise, or redundancies in the data?}

There are several potential sources of error and redundancy in the data:
\begin{itemize}
    \item \textit{Parsing issues}: Questions with options represented as tables might have been parsed incorrectly. Some parsing errors may result in typos (e.g., words broken into two segments) or missing words at the end of an option. Many of these errors have been manually corrected, especially in the test split, which should be free of such issues.
    \item \textit{Image noise}: The images for college admissions can present noise, but Gemini 1.5 Flash processed them relatively well. Some hallucinations may still exist, although we manually searched for them.
    \item \textit{Duplicates}: Some questions and options are duplicated across different problem sets or even within the same source. We have marked the obvious duplicates, but repetition of questions and answer options could still occur.
    \item \textit{Answer errors}: Some answers might be wrong due to parsing errors or LLM hallucinations. Although we have manually checked every parsed answer, human error is still a possibility. Additionally, there could be mistakes in the original answer sheets, where wrong answers may have been transcribed. Despite thorough checks (as the collected data is from national contests with official sources), it is possible that a few incorrect answers might have slipped through.
    \item \textit{Image dependent questions}: We have tried to filter out any question that was dependent on a figure, as we do not intend for the dataset at the moment to be multi-modal, but some questions might have slipped through. This is possible only for the olympiad questions.
\end{itemize}
\bigskip

\subsection{Data Pre-processing}
\bigskip

\textbf{What pre-processing/cleaning was done?}

After extraction, several pre-processing and cleaning steps were applied to standardize and structure the data: 

1. Extracted the question number from the question text and placed it in a separate field. 

2. Standardized option identifiers to uppercase letters.

3. Ensured all options followed the structure: "\texttt{[identifier]. [text]}", where \texttt{[identifier]} is either a letter (\textit{A–D}, or \textit{A-E} for five-option lists) or a number (\textit{1–4} for group-choice questions).

4. Replaced multiple spaces with a single space.

5. Replaced newline characters with spaces.

6. Standardized quotes by replacing Romanian quotation marks with English ones.

7. Normalized diacritics to proper Romanian characters (e.g., \texttt{ș, ț, â, ă}).

8. Manually corrected grammar issues and typos.

9. Removed trailing characters such as commas, dots, spaces, and semicolons from option texts. 

10. Made Gemini 1.5 Flash act as a grammar correcting tool to help us further find typos. Manually checked the output of it as the LLM has a tendency to replace words besides the typos. (Also used Gemma-2-9B when Gemini 1.5 Flash was unavailable).
\bigskip

\textbf{Was the “raw” data saved in addition to the preprocessed/cleaned data?}

The PDF files are saved privately.
\bigskip

\textbf{Is the pre-processing software available? }

No.
\bigskip

\textbf{Does this dataset collection/processing procedure achieve the motivation for creating the dataset stated in the first section of this datasheet?}

This dataset successfully provides specialized (Romanian) biology terms that can be used for training or knowledge evaluation.

\bigskip

\section{Prompts}
\label{sec:appendix_prompts}
\small 

\section*{User Prompts Used for Benchmarking}

\noindent
\textbf{Single Choice}  
\begin{quote}
\%question-text\%  

You received a biology question in Romanian with multiple options. The biology question is collected from either national high school olympiads or admission exams for medical universities. Only one answer is correct.  

You will output only the letter of the right answer. Do not give any explanations.  

The letter of the right answer is:
\end{quote}

\noindent
\textbf{Group Choice}  
\begin{quote}
\%question-text\%  

You received a biology question in Romanian with multiple numbered options. The question is from national high school olympiads or medical university admission exams.  

To answer:  \\
1. Identify correct options.  \\
2. If only option 4 is correct, the answer must be D.  \\
3. If only options 1,3 are correct, the answer must be B.  \\
4. If only options 2,4 are correct, the answer must be C.  \\
5. If only options 1,2,3 are correct, the answer must be A.  \\
6. If all options are correct, the answer must be E.  \\

Do not give any explanations.  

The right answer is:
\end{quote}

\noindent
\textbf{Multiple Choice}  
\begin{quote}
\%question-text\%  

You received a biology question in Romanian with multiple options. The question is from national high school olympiads or medical university admission exams. One or multiple answers are correct.  

You will output the letter(s) of all the correct answers. Do not give any explanations.  

The letters of the right answers, as compact as possible, are:
\end{quote}

\section*{System Prompts Used for Benchmarking}

We include only five-shot prompts; one- and three-shot follow the same format with fewer questions. The displayed prompts use translated questions, but LLMs receive the original Romanian versions.
\\

\noindent
\textbf{Single Choice - Five Shot}  
\begin{quote}
Here are some examples of biology questions in Romanian with multiple options and the correct format for answering them:  

\# Question: The prokaryotic cell:  \\
A. characterizes viruses, bacteria, and blue-green algae  \\
B. contains peptidoglycan in the composition of the cell membrane  \\
C. does not have a cell wall  \\
D. the nuclear material is a circular double-stranded DNA molecule  \\
\# Answer: D  \\
---  \\
\# Question: The mesosomes of prokaryotes:  \\
A. have a role in respiration  \\
B. are made up of rRNA and proteins  \\
C. are invaginations of the plasma membrane in the form of lamellae  \\
D. have a role in photosynthesis  \\
\# Answer: A  \\
---  \\
\# Question: The sciatic nerve:  \\
A. is a cranial nerve  \\
B. contains only motor fibers  \\
C. contains both sensory and motor fibers  \\
D. originates in the medulla oblongata  \\
\# Answer: C  \\
---  \\
\# Question: Contain hydrolytic enzymes with a role in intracellular digestion:  \\
A. ribosomes  \\
B. lysosomes  \\
C. centrosome  \\
D. centrioles  \\
\# Answer: B  \\
---  \\
\# Question: Photosynthetic plastids are:  \\
A. oleoplasts  \\
B. leucoplasts  \\
C. rhodoplasts  \\
D. amyloplasts  \\
\# Answer: C  \\
\end{quote}

\noindent
\textbf{Group Choice - Five Shot}  
\begin{quote}
Here are some examples of biology questions in Romanian with multiple numbered options and the correct format for answering them:  

\# Question: Organic substances with a structural role include:  \\
1. lipids  \\
2. carbohydrates  \\
3. proteins  \\
4. nucleic acids  \\
\# Explanation: 1,3 are correct; 2,4 are not  \\
\# Answer: B  \\
---  \\
\# Question: The fundamental substance is present in the structure of:  \\
1. mitochondria  \\
2. chloroplasts  \\
3. the nucleus  \\
4. vacuoles  \\
\# Explanation: 1,2,3 are correct; 4 is not  \\
\# Answer: A  \\
---  \\
\# Question: The nucleolus:  \\
1. is surrounded by its own membrane  \\
2. is the densest part of the nucleus  \\
3. is the site of mRNA synthesis  \\
4. its volume depends on the physiological state of the cell  \\
\# Explanation: 2,4 are correct; 1,3 are not  \\
\# Answer: C  \\
---  \\
\# Question: The granum of chloroplasts:  \\
1. is found freely in the stroma  \\
2. contains DNA, RNA, proteins, and metals  \\
3. is surrounded by a double porous membrane  \\
4. contains photosynthetic pigments  \\
\# Explanation: 4 is correct; 1,2,3 are not  \\
\# Answer: D  \\
---  
\# Question: The interphase:  \\
1. represents the time interval between two successive cell divisions  \\
2. is characterized by DNA, RNA, and protein synthesis  \\
3. is the most metabolically active stage  \\
4. precedes the division phase of the cell cycle  \\
\# Explanation: 1,2,3,4 are correct  \\
\# Answer: E  \\
\end{quote}

\noindent
\textbf{Multiple Choice - Five Shot}  
\begin{quote}
Here are some examples of biology questions in Romanian with multiple options and the correct format for answering them:  

\# Question: The heart:  \\
A. has the mitral valve between the right atrium and right ventricle  \\
B. is equipped with trabeculae in the atria  \\
C. is a parenchymatous organ due to its strong ventricular musculature  \\
D. is equipped with 2 valves  \\
E. contains the His bundle, which plays a role in automatism with a discharge frequency of 25 impulses/min  \\
\# Answer: E  \\
---  \\
\# Question: The right atrium is characterized by:  \\
A. containing the sinoatrial node  \\
B. having trabeculae inside  \\
C. receiving the inferior venae cavae  \\
D. having a systole duration of 0.1s  \\
E. being the site where pulmonary veins open  \\
\# Answer: ACD  \\
---  \\
\# Question: The following associations are correct:  \\
A. chordae tendineae - atrioventricular valves  \\
B. sinoatrial node - interatrial septum  \\
C. cardiac cycle - 0.8s at a heart rate of 100 beats/min  \\
D. venous pressure at the level of the right atrium is 10 mmHg  \\
E. tricuspid valve - right atrioventricular orifice  \\
\# Answer: AE  \\
---  \\
\# Question: Arteries that originate directly from the subclavian artery include:  \\
A. external carotid  \\
B. vertebral  \\
C. brachial  \\
D. internal thoracic  \\
E. anterior intercostal  \\
\# Answer: BD  \\
---  \\
\# Question: The pulmonary veins:  \\
A. are two in number  \\
B. open into the left atrium, which contains the sinoatrial node  \\
C. are part of the small circulation, which begins in the right ventricle  \\
D. bring oxygenated blood to the heart from the alveolar-capillary membrane, which has an average thickness of 0.6 microns  \\
E. like the venae cavae, bring venous blood into the atria  \\
\# Answer: CD  \\
\end{quote}

\end{document}